\documentclass[11pt,a4paper]{article}
\usepackage[hyperref]{acl2017}
\usepackage{times}
\usepackage{latexsym}
\usepackage[inline]{enumitem}
\usepackage[T1]{fontenc}
\usepackage{url}
\usepackage{multirow}
\usepackage{csquotes}
\usepackage{microtype}
\usepackage{hyperref}
\usepackage{booktabs}
\usepackage{amsmath}

\aclfinalcopy 

\newcommand*{\affmark}[1][*]{\textsuperscript{#1}}

\title{Idea density for predicting Alzheimer's disease from transcribed speech}

\author{Kairit Sirts\affmark[1], Olivier Piguet\affmark[2,3] and Mark Johnson\affmark[4] \\\\
\affmark[1]Institute of Computer Science, University of Tartu \\
\affmark[2]School of Psychology and Brain \& Mind Centre, The University of Sydney \\
\affmark[3]Neuroscience Research Australia, The University of New South Wales \\
\affmark[4]Department of Computing, Macquarie University \\
 {\tt kairit.sirts@ut.ee}, {\tt olivier.piguet@sydney.edu.au} \\
 {\tt mark.johnson@mq.edu.au}}

\date{}

\begin{document}
\maketitle
\begin{abstract}
Idea Density (ID) measures the rate at which ideas or elementary predications are expressed in an utterance or in a text.
Lower ID is found to be associated with an increased risk of developing Alzheimer's disease (AD) \cite{Snowdon1996,Engelman2010}.
ID has been used in two different versions: propositional idea density (PID) counts the expressed ideas and can be applied to any text while semantic idea density (SID) counts pre-defined information content units and is naturally more applicable to normative domains, such as picture description tasks.
In this paper, we develop DEPID, a novel dependency-based method for computing PID, and its version DEPID-R that enables to exclude repeating ideas---a feature characteristic to AD speech.  We conduct the first comparison of automatically extracted PID and SID in the diagnostic classification task on two different AD datasets covering both closed-topic and free-recall domains. 
While SID performs better on the normative dataset, adding PID leads to a small but significant improvement (+1.7 F-score). On the free-topic dataset, PID performs better than SID as expected (77.6 vs 72.3 in F-score) but adding the features derived from the word embedding clustering underlying the automatic SID increases the results considerably, leading to an F-score of 84.8.
\end{abstract}

\section{Introduction}

Idea density (ID) measures the rate of propositions or ideas expressed per word in a text and it is connected to some very interesting results from neuroscience related to Alzheimer's disease (AD). In particular, two longitudinal studies---the Nun Study \cite{Snowdon1996} and the Precursors Study \cite{Engelman2010}---suggest that lower ID, as measured from the essays written in young age, is associated with the higher probability of developing AD in later life.

  \begin{table}
  \begin{small}
 \setlength\tabcolsep{5.5pt}
 \centering
 \begin{tabular}{l|l}
 \toprule
  \multicolumn{2}{c}{\textbf{The old gray [MARE] has a very large [NOSE].}}\\
  \midrule
  \bf Dependencies & \bf Propositions \\
  \midrule
  det(The, mare) & \\
  amod(old, mare) & (\textsc{old}, \textsc{mare})\\
  amod(gray, mare) & (\textsc{gray}, \textsc{mare}) \\
  nsubj(mare, has) & (\textsc{has}, \textsc{mare}, \textsc{nose})\\
  det(a, nose) & \\
  advmod(very, large) & (\textsc{very}, (\textsc{large}, \textsc{nose}))\\
  amod(large, nose) & (\textsc{large}, \textsc{nose})\\
  dobj(nose, has) & (\textsc{has}, \textsc{mare}, \textsc{nose})\\
  punct(., has) & \\
 \end{tabular}
 \caption{The alignment of the dependency and propositional structures. 
 The example sentence is due to \protect\newcite{Brown2008}. The predicative proposition (\textsc{has}, \textsc{mare}, \textsc{nose}) is represented by two dependency arcs.}
 \label{fig:pid_and_dep}
  \end{small}
\end{table}

Two alternative definitions of idea density have been used in relation to AD. 
\textbf{Propositional idea density} (PID) counts the number of \emph{any} ideas expressed in the text, setting no restriction to the topic \cite{Turner1977,Chand2010}. An example sentence with its ideas or propositions is given in Table~\ref{fig:pid_and_dep}. Based on each proposition a question can be formulated with a yes or no answer. Removing a proposition from a sentence changes the semantic meaning of that sentence. For instance, removing the proposition (\textsc{gray}, \textsc{mare}) from the example makes the overall meaning of the sentence more general. The PID is then computed by normalising the proposition count with the token count and thus the PID of the example given in Table~\ref{fig:pid_and_dep} is $6 / 9 \approx 0.667$. 

The existing tool for automatic PID computation, CPIDR \cite{Brown2008}, is based on counting POS tags. 
However, we noticed that the propositional structure of a sentence  is very similar to its dependency structure, see the first column in Table~\ref{fig:pid_and_dep}.
This motivated us to come up with DEPID, a method for computing PID from dependency structures. In addition, DEPID more easily enables to consider idea repetition which has been shown to be a characteristic feature in Alzheimer's speech \cite{Bayles1985,Tomoeda1996,Bayles2004}, resulting in a modified PID version DEPID-R which excludes the repeated ideas.

\textbf{Semantic idea density} (SID) \cite{Ahmed2013a,Ahmed2013} relies on a set of pre-defined information content units (ICU).
ICU is an object or action that can be seen on the picture or is told in the story and is expected to be mentioned in the narrative. For instance, assuming that the words in capital letters and square brackets in the example sentence shown in Table~\ref{fig:pid_and_dep} belong to the set of pre-defined ICUs the SID is computed by normalising the ICU count with the token count: $2 / 9 \approx 0.222$. 
Recently, \newcite{Yancheva2016}, proposed a method for computing SID based on word embedding  clusters. We use their method for computing SID as it does not rely on any pre-defined ICU inventory and thus is applicable also on free-topic datasets.

PID and SID are complementary definitions of idea density with SID being naturally applicable in standardised picture description or story re-telling tasks while PID is more suitable on datasets of spontaneous speech on free topics.

In this paper we study the predictiveness of both PID and SID features in the diagnostic classification task for predicting AD. To that end, we conduct experiments on two very different datasets: DementiaBank, which consists of transcriptions of a normative picture description task,  and AMI, which contains autobiographical memory interviews describing life events freely chosen by the subjects.

We show that on the DementiaBank data the POS-based PID scores are actually higher for AD patients than they are for normal controls, contrary to the expectations from the AD literature \cite{Engelman2010,Chand2012,Kemper2001}. By studying the characteristics of the DementiaBank we are able to adapt DEPID such that its PID values become significantly different between the patient and control groups in the expected direction. Thus, we believe that our proposed DEPID is a better tool for measuring PID as described by neurolinguists on spontaneous speech transcripts than the POS-based CPIDR.

Secondly, we show that the SID performs better than PID on the constrained-domain DementiaBank corpus but adding the PID feature leads to a small but significant improvement.

Thirdly, we show that on the free-topic AMI dataset the PID performs better than the automatically extracted SID, but adding the features derived from the word embedding clustering underlying the SID, modeling the broad discussion topics, increases the results considerably---an effect which is less visible on the constrained topic DementiaBank.

The contributions of this paper are the following:
\begin{enumerate}[noitemsep]
\item Development of DEPID, the new dependency-based method for automatically computing PID and its version DEPID-R which enables to detect and exclude idea repetitions;
\item Analysis of the characteristic features of the DementiaBank dataset and the proposal for modifying DEPID to make it applicable to this and other similar closed-topic datasets. 
\item Results of extensive diagnostic classification experiments using PID, SID and several related baselines on two very different AD datasets.
\end{enumerate}

\section{Idea density and Alzheimer's disease}

ID was first associated with AD in the Nun Study \cite{Snowdon1996}, based on a cohort of elderly nuns participating in a longitudinal study of aging and Alzheimer's disease. In this work, they studied the autobiographical essays the nuns had written decades ago in their youth.
The nuns were divided into three groups based on their ID score computed from the essays, so that each group covered 33.3\% percentile of the whole range of ID values. The lowest group was labeled as having low ID and the medium and highest group as having high ID. These groups were established from a sample of 93 nuns.
The association between AD and ID was studied on a sample of 25 nuns who had died by the time of the study, for 10 of whom the cause of death had been marked as AD. 
The study found that most subjects with AD
belonged to the low ID group while most of those, who did not develop AD, belonged to the group with high ID, thus suggesting that the low ID in youth might be associated with the development of the AD in later life.

Similar work was conducted on a group of medical students for whom essays from the time of their admission to the medical school several decades earlier were available \cite{Engelman2010}. 
The results of this study also showed a significantly lower ID on the AD group as compared to the healthy controls, suggesting that ID could be an important discriminative feature for predicting AD.

\subsection{Propositional and semantic idea density}

Two different versions of ID have been developed over time, both derived from the propositional base structure developed by \newcite{Kintsch1973}  to describe the semantic complexity of texts in reading experiments.  

\emph{Propositional idea density} (PID), which was used both in the Nun Study and the medical students study, is based on counting the semantic propositions as defined by \newcite{Turner1977} and later refined by \newcite{Chand2012}. Three main types of propositions where described: 
\begin{enumerate*}[label=\arabic*)]
\item predications that are based on verb frames;
\item modifications that include all sorts of modifiers, e.g. adjectival, adverbial, quantifying, qualifying etc.; and
\item connections that join simple propositions into complex ones.
\end{enumerate*}
For each proposition, a question can be formed with a yes or no answer. For instance, based on the example in Table~\ref{fig:pid_and_dep}, we could form the following questions:
\begin{enumerate}[topsep=2pt]
\itemsep -0.5em
\item Is the mare old?
\item Is the mare gray?
\item Has the mare a nose?
\item Is the nose large?
\item Is the nose very large?
\end{enumerate}

Each of those questions inquires about a different aspect of the whole sentence and is a basis of an idea or proposition.

\emph{Semantic idea density} (SID) has retained its relation to the propositional base of some text. It relies on a set of information content units (ICUs) that have been pre-defined for a closed-topic task, such as picture description or story re-telling. For instance, different inventories of 7-25 ICUs have been described for the Cookie Theft picture task \cite{Goodglass1983}, listing objects visible on the picture such as \emph{\enquote{boy}}, \emph{\enquote{girl}}, \emph{\enquote{cookie}} or \emph{\enquote{kitchen}} or actions performed on the scene such as \emph{\enquote{boy stealing cookies}} or \emph{\enquote{woman drying dishes}}. 
SID is computed by counting the number of ICUs mentioned in the text and then normalising by the total number of word tokens.

\subsection{Related work on AD using ID}
PID, computed with CPIDR, has been used in few previous works for predicting AD. \newcite{Jarrold2010} used PID as one among many features and reported it as significant. They obtained a classification accuracy of  73\% on their dataset, which contained short structured clinical interviews, with their best model and feature set that also included the PID feature.  
PID was also used by \newcite{Roark2011} to detect mild cognitive impairment on a story re-telling dataset. However, they found no significant difference between groups in terms of PID and thus, their feature selection procedure most probably filtered it out.

In terms of SID, most previous work has relied on manually defined ICUs \citep{Ahmed2013,Ahmed2013a}. 
\newcite{Fraser2015} extracted binary and frequency-based ICU features. They searched for words related to the ICU objects and looked at the \emph{nsubj}-relations in the dependency parses to detect the ICUs referring to actions. The binary feature was set when any word related to an ICU was mentioned in the text, while frequency-based features counted the total number of times any word referring to an ICU was mentioned. 

Recently, \newcite{Yancheva2016} proposed a method for automatically extracting ICUs and computing SID without relying on a manually defined ICU inventory. This work will be reviewed in more detail in section~\ref{sec:sid}. They found that the automatically extracted ICUs and SID performed as well in a diagnostic AD classification task as the human-defined ICUs.

\section{Computation of PID}
\label{sec:depid}

Automating the computation of PID is difficult because it is essentially a semantic measure. The instructions given by \newcite{Turner1977} for counting the propositions assume the comprehension of the semantic meaning of the text, while the raw text lacks the necessary semantic annotations.
However, it has been noticed that the propositions roughly correspond to certain POS tags. In particular, \newcite{Snowdon1996} mention that elementary propositions are expressed using verbs, adjectives, adverbs and prepositions. This observation is the basis of the CPIDR program \cite{Brown2008}, a tool for automatically computing PID scores from text. CPIDR first processes the text with a POS-tagger, then counts all verbs, adjectives, adverbs, prepositions and coordinating conjunctions as propositions, and then applies a set of 37 rules to adjust the final proposition count. 

\begin{table}[t]
\begin{minipage}{0.45\textwidth}
\begin{small}
\centering
\begin{tabular}{l|l}
\bf Dep rel & \bf Proposition type \\
\hline
advcl & Causal connection \\
advmod & Qualifying modification \\
amod & Qualifying modification \\
appos & Referencial predication \\
cc & Conjunctive connective \\
csubj & Predication with a clausal subject \\
csubjpass & Predication with a passive clausal \\
& subject \\
det\footnote{except \emph{a}, \emph{an} and \emph{the}} & Quantifying modification \\
neg & Negative modification \\
npadvmod & Qualifying modification \\
nsubj\footnote{except \emph{it} and \emph{this}} & Predication subject \\
nsubjpass & Predication with passive subject \\
nummod & Quantifying modification \\
poss & Possessive modification \\
predet & Qualifying modification \\
preconj & Conjunctive or disjunctive \\
& connection \\
prep & Proposition denoting purpose, \\ 
& location, intention, etc.\\
quantmod & Quantifying modification \\
tmod & Qualifying modification \\
vmod & Qualifying modification \\
\end{tabular}
\caption{Dependency relations encoding propositions.}
\label{tab:dep_manual}
\end{small}
\end{minipage}
\end{table}

\subsection{DEPID---dependency-based PID}

We propose that the dependency structure is better suited for PID computation than the POS tag counting approach adopted by the existing CPIDR program \cite{Brown2008} because the dependency structure resembles more closely the semantic propositional structure, see Table~\ref{fig:pid_and_dep}.
We treat each dependency type as a separate feature and manually set the feature weights to either one or zero depending on whether this dependency relation encodes a proposition or not.
We make these decisions based on the dependency type descriptions in the Stanford dependency manual \cite{Marneffe2008}.
The dependency types with non-zero weights are listed in Table~\ref{tab:dep_manual}. The PID is then computed by summing the counts of those dependency relations
and normalising by the number of word tokens. We call our dependency-based PID computation method DEPID.

We computed the Spearman correlations between CPIDR, DEPID and manual proposition counts on the 69 example sentences given in chapter 2 in \cite{Turner1977}\footnote{Similar to \newcite{Brown2008}, we exclude the example 17, but for examples 18, 54, 55, 56, we include all paraphrases.} and the 177 example sentences given in \cite{Chand2010}, making up the total of 276 sentences. These correlations are given in Table~\ref{tab:prop_corr}. We observe that by just counting the dependency relations given in Table~\ref{tab:dep_manual}, we obtain proposition counts that correlate
better with the manual counts than the POS-based CPIDR counts.  

\begin{table}[t]
 \centering
 \begin{tabular}{lc}
 & \bf Spearman r \\
 \toprule
  CPIDR vs Manual & 0.795 \\
  DEPID vs Manual & 0.839 \\
  DEPID vs CPIDR & 0.864 \\ 
 \end{tabular}
 \caption{Spearman correlations between CPIDR, DEPID and manual proposition counts on the examples given in \protect\newcite{Turner1977} and \protect\newcite{Chand2010}.}
 \label{tab:prop_corr}
\end{table}

\subsection{DEPID-R}
It is known that the Alzheimer's language is generally fluent and grammatical but in order to maintain the fluency the deficiencies in semantic or episodic memory are compensated with empty speech \cite{Nicholas1985}, such as repetitions, both on the word level but also on the idea, sentence or narrative level. 
DEPID easily enables to track repeated ideas in the narrative. We consider a proposition as repetition of a previous idea when the \emph{deprel}(\textsc{dependent lemma},  \textsc{head lemma}) tuples of the two propositions match. For instance, a sentence \emph{\enquote{I had a happy life.}} contains three propositions: \emph{nsubj}(\textsc{I}, \textsc{have}), \emph{dobj}(\textsc{life}, \textsc{have}) and \emph{amod}(\textsc{happy}, \textsc{life}). Another sentence \emph{\enquote{I've had a very happy life.}} later in the same narrative only adds a single proposition to the total count---\emph{advmod}(\textsc{very}, \textsc{happy})---as this is the only new piece of information that was added. 

We modify DEPID to exclude the repetitive ideas of a narrative by only counting the proposition \emph{types} expressed with the lexicalised \emph{deprel}(\textsc{dependent lemma},  \textsc{head lemma}) dependency arcs.
We call this modified version of dependency-based PID computation method DEPID-R. The relation between DEPID-R and DEPID is that DEPID counts the \emph{tokens} of the same propositions.

\section{Computation of SID}
\label{sec:sid}

Recently, \newcite{Yancheva2016} proposed a method for automatically computing SID without the use of manually defined ICUs. Their method relies on clustering word embeddings of the nouns and verbs found in the transcriptions, assuming that the embeddings of the words related to the same semantic unit are clustered together. 

They first perform K-means clustering on the word embeddings. Then, for each cluster they compute the mean distance $\mu_{cl}$ and its standard deviation $\sigma_{cl}$. The mean distance is the average Euclidean distance of all vectors assigned to a cluster from the  centroid of that cluster. Finally, for each word they compute the scaled distance as a z-score of the Euclidean distance  $d_E$ between the word embedding and its closest cluster centroid:
\begin{equation*}
d_{scaled} = \frac{d_{E}- \mu_{cl}}{\sigma_{cl}}
\end{equation*}
The words with $d_{scaled} < 3$ are counted as automatic ICUs. SID is then computed by dividing the number of ICUs with the total number of word tokens in the transcription.

In addition to SID, \newcite{Yancheva2016} experiment with distance-based features also derived from the same clustering. The distance feature for each cluster is computed as the average of the scaled distances of the words (nouns or verbs) in the transcript assigned to that cluster. 
These cluster features are not directly related to the concept of SID but they could be viewed as an automatic approximation of features derived from the human annotated ICUs.

\section{Experiments}

\subsection{Data}
\label{sec:data}

\begin{table}[t]
\setlength\tabcolsep{6pt}
\centering
\begin{tabular}{lllll}
& \multicolumn{2}{c}{\bf DB} & \multicolumn{2}{c}{\bf AMI} \\
 & \bf AD & \bf Ctrl & \bf AD & \bf Ctrl \\
\toprule
Subjects & 169 & 98 & 20 & 20\\
Samples & 257 & 241 & 36 & 20\\
Mean samples & 1.52 & 2.46 & 1.80 & 1.00 \\
Mean words & 104 & 114 & 1674 & 1509\\
Std words & 58 & 59 & 778 & 688 \\
\end{tabular}

\caption{Statistics of the DementiaBank (DB) and AMI datasets. \emph{Mean samples} is the average number of samples per subject. \emph{Mean} and \emph{std words} are the mean number of words per sample and the respective standard deviation.}
\label{tab:data_stat}
\end{table}

We conduct experiments on two very different AD datasets. The first dataset is derived from the DementiaBank \cite{Becker1994}, which is part of a publicly available Talkbank corpus.\footnote{\url{https://talkbank.org/DementiaBank/}} It contains descriptions of the Cookie Theft picture \cite{Goodglass1983} produced by subjects diagnosed with dementia as well as of healthy control cases.  
The data is manually transcribed and annotated in the CHAT format \cite{MacWhinney2000}, containing a range of annotations denoting various speech events. This is the same dataset used by \newcite{Yancheva2016} and similar to them, we use the interviews of all control subjects and subjects whose diagnose is either AD or probable AD.  

The second dataset, collected at NeuRA\footnote{Neuroscience Research Australia}, contains autobiographical memory interviews (AMI) of both AD patients and healthy control subjects. Each interview consists of four stories, each story describing events from a particular period of the subject's life: teenage years, early adulthood, middle adulthood and last year. Each story has three logical parts: free recall, general probe and specific probe. In the free recall part the subject is asked to talk freely about events he remembers from the given life period. 
In the general recall part the interviewer helps to narrow down to a particular specific event. 
In the specific probe part the interviewer asks a number of predefined questions about this specific event. We use all four stories of an interview as a single sample but extract only the free recall part of each story as this is the most spontaneous part of the interview. 

We preprocess both data sets similarly, following the procedure described in \cite{Fraser2015} as closely as possible. We first extract only the patient's dialogue turns. Then we remove any tokens that are not words (e.g. laughs). In DementiaBank corpus, such tokens can be detected by various CHAT annotations. We also remove filled pauses such as \emph{um}, \emph{uh}, \emph{er}, \emph{ah}. The statistics of both datasets are given in Table~\ref{tab:data_stat}.

\subsection{Analysis of the idea density}
\label{sec:pid_ana}

\begin{table}[t]
\setlength\tabcolsep{3.0pt}
\begin{small}
\centering
\begin{tabular}{llcc}
\bf Data & \bf Method & \bf AD mean (sd)  & \bf Ctrl mean (sd) \\
\toprule
DB & CPIDR* & 0.518 (0.069) & 0.491 (0.057) \\
DB & DEPID* & 0.371 (0.052) & 0.356 (0.046) \\ 
DB & DEPID-R & 0.339 (0.049) & 0.334 (0.042)\\
DB & DEPID-R-ADD* & 0.168 (0.064) & 0.194 (0.059) \\
DB & SID* & 0.380 (0.051) & 0.427 (0.045) \\ 
\midrule
AMI & CPIDR & 0.524 (0.023) & 0.532 (0.017) \\
AMI & DEPID & 0.468 (0.022) & 0.473 (0.017) \\
AMI & DEPID-R* & 0.334 (0.027) & 0.366 (0.027) \\
AMI & DEPID-R-ADD+* & 0.291 (0.032) & 0.337 (0.032) \\
AMI & SID* & 0.346 (0.034) & 0.385 (0.024)
\end{tabular}
\caption{The statistics of the ID values for AD and control groups. DEPID-R ignores the repeated ideas. DEPID-R-ADD for DementiaBank additionally excludes conjunctions, sentences with \emph{I} and \emph{you} subjects and sentences with vague meaning. DEPID-R-ADD+ for AMI only ignores sentences with vague meaning. SID is computed based on the clustering of the whole dataset. Star (*) after the method name indicates that the difference in group means is statistically significant ($p<0.001$).}
\label{tab:pid_stat}
\end{small}
\end{table}

First, we perform a statistical analysis of the different ID measures in Table~\ref{tab:pid_stat} on both datasets using the indepedent samples Wilcoxon rank-sum test to test the difference between group means. 

The DEPID computed PID values are systematically lower than the CPIDR values on both datasets, suggesting that either CPIDR overestimates or the DEPID  
underestimates the number of propositions. In order to check that we manually annotated the propositions of 20 interviews from DementiaBank according to the guidelines given by \newcite{Chand2012}.
We found that both CPIDR and DEPID overestimate the PID values although CPIDR does it to much greater extent.  CPIDR both overestimates the number of propositions and underestimates the number of tokens in certain cases leading to higher PID scores. For example, CPIDR does not count contracted forms, such as \emph{\enquote{'s}} in \emph{\enquote{it's}} or \emph{\enquote{n't}} in \emph{\enquote{don't}} as distinct tokens. Because there are many such forms in DementiaBank transcriptions, this behaviour considerably lowers CPIDR token counts. Also, CPIDR counts each auxiliary verb in present participle constructions as a separate proposition although these auxiliaries only mark syntax, thus leading to an artificially high proposition count. For instance, the clauses \emph{\enquote{she is reaching}} and \emph{\enquote{he is taking}} both contain two propositions according to CPIDR, whereas they both really contain only one semantic idea.

Both CPIDR and DEPID PID values differ significantly between AD and control groups on DementiaBank but the mean values are opposite to what was expected---AD patients have significantly higher PID than controls.  
When the repeated ideas are not counted (DEPID-R), the difference between groups becomes non-significant.
However, we were curious about why the association between the lower PID values and the AD diagnosis cannot be observed on DementiaBank.
Thus, we investigated this issue and found that the DementiaBank interviews have certain additional characteristics that contribute to the automatic proposition count being too high. 

\paragraph{Conjunctive propositions} 
First, we noticed that most \emph{and-conjunctions} are used as lexical fillers in DementiaBank, whereas both CPIDR and DEPID count all conjunctions as propositions. In order to address this problem we excluded the \emph{cc} dependency type from the set of propositions.

 \paragraph{Sentences with pronominal subjects} 
Secondly, we noticed that the sentences with subject either \emph{I} or \emph{you} most probably do not say anything about the picture but rather belong to the meta conversation.
Two examples of such sentences are for instance \emph{\enquote{what else can I tell you about the picture?}} or \emph{\enquote{I'd say that's about all.}}. To solve this problem we did not count propositions from sentences, where  the subject was either \emph{I} or \emph{you}.

\paragraph{Vague sentences}
Finally, we observed that the AD patients seem to utter more \emph{vague sentences} that do not contain any concrete ideas, such as for instance \emph{``the upper one is there''} or \emph{``they're doing more things on the outside.''}. Both CPIDR and DEPID extract propositions from syntactic structures and thus they count pseudo-ideas from those sentences as well. 
To detect such vague sentences we evaluated the specificity of all sentences using SpeciTeller \cite{Li2015}. SpeciTeller predicts a specificity score between 0 and 1 for each sentence using features extracted from the sentence surface-level, specific dictionaries and distributional word embeddings.
We did not count propositions from sentences whose specificity score was lower than 0.01. 

After incorporating all those three measures to DEPID  we finally obtain PID values on DementiaBank  that are significantly different for patients and controls in the expected direction---the AD patients have significantly lower PID values than control subjects. Note that those measures only affect the proposition count and not the number of tokens. Also note that although these measures were motivated by the observations made on one particular (DementiaBank) dataset, they can be expected to be applicable to other similar closed-topic datasets, containing picture descriptions or story re-tellings.\footnote{Unfortunately, aside from DementiaBank there are no other publicly available AD datasets and thus we could not test whether our expectations hold true.}

On AMI data, the difference between group means is non-significant for both CPIDR and DEPID values. However, when the repeated ideas are excluded (DEPID-R), the mean PID for AD patients is significantly lower than for controls, as expected.
It should be noted that the first two problems observed on DementiaBank---conjunctions and pronominal subjects---are not actual on the free-recall AMI data.
In autobiographical memory interviews many sentences are expected to have \emph{I} as subject. Also, the \emph{and}-conjunctions are more likely to convey real ideas there rather than carry the role of lexical fillers. However, AD patients can utter more sentences with very vague meaning in AMI data as well and thus, 
in the last row of the Table~\ref{tab:pid_stat} we show the DEPID PID values with vague sentences excluded for AMI dataset as well. We see that the PID values decrease for both patients and controls and the difference between groups remains statistically significant.

SID values differ significantly between the AD and control groups on both datasets with AD patients having significantly lower SID values as expected. The clustering underlying the automatically computed SID is trained on the whole dataset for both DementiaBank and AMI data.

\subsection{Classification setup}

We test both PID and SID  in the diagnostic binary classification task on both DementiaBank and AMI datasets. When computing PID, the repeated ideas are excluded (DEPID-R). In addition, for DementiaBank, we also use the additional measures described in Section~\ref{sec:pid_ana} (DEPID-R-ADD) as, according to Table~\ref{tab:pid_stat}, just DEPID-R cannot be expected to be predictive on that type of dataset.
We compute the SID as described in Section~\ref{sec:sid}. 
In following \cite{Yancheva2016}, we cluster the 50-dimensional Glove embeddings\footnote{\url{http://nlp.stanford.edu/projects/glove/}} of all nouns and verbs found in the transcripts with k-means.
Similar to them, we set the number of clusters to 10 on both datasets.

For single feature models (SID or PID) we use a simple logistic regression classifier. For models with multiple features we use the elastic net logistic regression with an elastic net hyperparameter $\alpha=0.5$. We train and test with 10-fold cross-validation on subjects and 
 repeat each experiment 100 times. We report the mean and standard deviation of the 100 macro-averaged cross-validated runs.
 For each experiment we report class-weighted precision, recall and F-score.\footnote{Classification accuracy is omitted because it is equivalent to the class-weighted recall.}

\subsection{Classification results}
\label{sec:classification}

\begin{table}[t]
\begin{small}
\centering
\begin{tabular}{llccc}
\bf Data & \bf Features & \bf Precision & \bf Recall & \bf F-score \\
\toprule
DB & CPIDR & 59.8 (0.7) & 59.1 (0.5) & 58.8 (0.5) \\
DB & PID & 61.1 (0.7) & 60.3 (0.6) & 60.0 (0.5) \\
DB & SID & 71.4 (0.6) & 70.7 (0.5) & 70.5 (0.5) \\
DB & SID+PID & {\bf 73.7} (0.9) & {\bf 72.1} (0.6) & {\bf 72.2} (0.6)\\
\midrule
AMI & CPIDR & 45.1 (3.2) & 63.4 (1.8) & 51.9 (2.3)\\
AMI & PID & 79.2 (1.9) & {\bf 80.0} (0.5) & 77.6 (0.9)\\
AMI & SID & 73.7 (3.0) & 75.3 (1.5) & 72.3 (2.1)  \\
AMI & SID+PID & {\bf 82.9} (3.8) & 78.0 (1.8) & {\bf 77.7} (1.8) \\
\end{tabular}
\end{small}
\caption{Classification results of various ID measures. The PID is DEPID-R-ADD for DementiaBank and DEPID-R for AMI.}
\label{tab:pid_results}
\end{table}

The classification results using various ID measures are shown in Table~\ref{tab:pid_results}. On both datasets, PID and SID are better from the CPIDR baseline although the difference is considerably larger on the free-recall AMI dataset. On DementiaBank, SID performs better than PID and combining SID and PID also gives a small consistent cumulative effect, improving the F-score by 1.7\%. On AMI data, the SID performs surprisingly well, considering that the automatic ICUs were  extracted from only 10 clusters and the number of clusters was not tuned to that dataset at all. However, PID performs ca 5\% better than SID in terms of all measures. Combining PID and SID gives some improvements in precision at the cost the decrease in recall and gives no cumulative gains in F-score. These results are fully in line with our expectations that the syntax-based DEPID performs better on the free-topic dataset, while the SID is better on closed-domain dataset.

\begin{table}[t]
\setlength\tabcolsep{4.9pt}
\begin{small}
\centering
\begin{tabular}{ll|ccc}
\bf Data & \bf Features & \bf Precision & \bf Recall & \bf F-score\\  
\toprule
DB & Clusters & 62.3 (1.6) & 62.2 (1.7) & 62.2 (1.7) \\
DB & C+PID & 67.4 (1.7) & 64.9 (1.5) & 65.1 (1.5)  \\
DB & C+SID & 73.4 (1.4) & 71.5 (1.3) & 71.6 (1.3)  \\
DB & C+SID+PID & {74.4} (1.5) & {72.5} (1.2) & {72.7} (1.2)  \\
DB & LIWC & 80.0 (0.9) & 78.4 (0.7) & 78.5 (0.7)\\ 
DB & BOW & {\bf 80.6} (1.1) & {\bf 79.1} (1.0) & {\bf 79.3} (1.0)  \\ 
\midrule
AMI & Clusters & 76.9 (7.7) & 71.2 (5.2) & 70.5 (5.8)  \\
AMI & C+PID & 81.2 (5.0) & 75.7 (3.8) & 75.3 (3.8)  \\
AMI & C+SID & 83.5 (5.0) & 77.9 (4.1) & 77.7 (4.4) \\
AMI & C+SID+PID & {\bf 84.6} (4.4) & {\bf 78.1} (3.8) & {\bf 78.4} (4.0)\\
AMI & LIWC & 74.2 (4.7) & 67.8 (3.5) & 66.8 (3.3) \\ 
AMI & BOW & 65.1 (7.2) & 65.3 (4.1) & 61.6 (4.7) \\
\end{tabular}
\caption{Classification results on DementiaBank (DB) and AMI using cluster features (C) combined with PID and SID, and LIWC and BOW baselines. The PID is DEPID-R-ADD for DementiaBank and DEPID-R for AMI.}
\label{tab:results}
\end{small}
\end{table}

\begin{table}[t]
\setlength\tabcolsep{4.9pt}
\begin{small}
\centering
\begin{tabular}{ll|ccc}
\bf Data &\bf Features&  \bf Precision & \bf Recall & \bf F-score \\  
\toprule
DB & Clusters &  68.0 (1.2) & 65.5 (0.9) & 65.7 (0.8) \\
DB & C+PID &  69.6 (1.1) & 67.1 (0.7) & 67.4 (0.7) \\
DB & C+SID & 75.3 (1.0) & 73.3 (0.7) & 73.5 (0.7) \\
DB & C+SID+PID & {\bf 76.6} (1.1) & {\bf 74.8} (0.8) & {\bf 75.0} (0.7) \\
\midrule
AMI & Clusters &  86.0 (3.6) & 80.4 (2.2) & 80.5 (2.1) \\
AMI & C+PID &  88.4 (3.9) & 83.0 (2.7) & 83.2 (2.8) \\
AMI & C+SID &  {\bf 88.6} (3.0) & {\bf  84.8} (1.7) & {\bf 84.8} (1.7) \\
AMI & C+SID+PID &  {87.3} (3.8) &{82.4} (2.6) & {82.7} (2.7) \\
\end{tabular}
\caption{Classification results on DementiaBank (DB) and AMI using cluster features (C) combined with PID and SID. The clusters are pre-trained on the whole dataset. The PID is DEPID-R-ADD for DementiaBank and DEPID-R for AMI.}
\label{tab:full_clusters}
\end{small}
\end{table}

For better comparison with \newcite{Yancheva2016} we also experimented with the distance-based cluster features, which are derived from the clusters underlying the automatic SID (see section~\ref{sec:sid}). We also show additional semantic baselines using LIWC features \cite{Tausczik2010} and bag-of-word (BOW) features extracting the counts of nouns and verbs normalised by the number of tokens. These results are shown in Table~\ref{tab:results}. 
On DementiaBank dataset, cluster features alone do not perform too well and using cluster features together with PID and SID gives only minor improvements. 
On the other hand, both the LIWC and BOW baselines perform very well on DementiaBank with BOW features giving the total highest precision of 80.6\%, recall of 79.1\% and F-score of 79.3\%. In fact, these results are very close to the state-of-the-art on this dataset: a recall of 81.9\% \citep{Fraser2015} and an F-score of 80.0\% \cite{Yancheva2016}. Note however that the BOW features are conceptually much simpler than the acoustic and lexicosyntactic features extracted by \newcite{Yancheva2016} and \newcite{Fraser2015}. 

On the free-recall AMI data, the cluster features perform surprisingly well while the results of the LIWC and BOW baselines are lower. Adding cluster features to ID behaves inconsistently---in case of SID the F-score improves while adding cluster features to PID lowers the F-score. It is also worth noticing that results on AMI data including cluster features vary quite a bit, in some cases having standard deviation even as high as 7.7\%.

Finally, we experimented with a scenario where the word embedding clusters are pre-trained on the whole dataset, in which case the clustering and thus also the SID feature reflect the structure of both training and test folds. This scenario assumes re-training the clustering and the classification model for each new test item/set. Although the classification model is then informed by the test set, we do not see it as test set leakage as the clustering is unsupervised. These results, given in Table~\ref{tab:full_clusters}, show that all results on both datasets improve, whereas the improvements are considerably larger on AMI dataset, which is expected because the model trained on the free-topic AMI data is likely to gain more on knowing the topics discussed in the test item/set. This scenario gives the highest  F-score of 84.8\% on this dataset when adding cluster features to SID.

Note, that the cluster features F-score trained on the full dataset is slightly lower than the 68\% reported by \newcite{Yancheva2016}. This difference is probably due to the differences in hyperparameters  and experimental setup: we use an elastic-net regularised logistic regression classifier while they used a random forest, we perform 10-fold cross-validation while they divided the DementiaBank into 60-20-20 train-dev-test partitions. However, the classification performance of cluster features together with SID are in the same range as their reported 74\%.

\section{Discussion}

This is the first work we are aware of that compares the same methods for predicting AD on two different datasets. 
Moreover, most previous work has been conducted either on constrained-topic datasets, containing picture descriptions \cite{Orimaye2014,Fraser2015,Yancheva2016,Rentoumi2014}, or semi-constrained structured interviews about some particular topic \cite{Thomas2005,Jarrold2010,Jarrold2014}, while our AMI dataset contains free recall samples and thus is probably more spontaneous than the previously used datasets.

We expected PID to perform well on the free-recall AMI dataset, which proved to be the case. However, we were surprised that the small number of automatically extracted clusters perform so well on that dataset too. 
This raises the natural question what topics those clusters represent. To shed light on this question, we studied the clustering trained on the whole AMI dataset. There were three clusters for which values differed significantly\footnote{We used the Wilcoxon signed rank test.} between AD and control subjects: C0 ($p<0.001$), C6 ($p < 0.001$) and C9 ($p=0.0044$). C0, which could be denoted as a cluster describing experiences, contained a diverse mix of words, which close to the cluster center denoted specific aspects of something or connoted emotions such as \emph{\enquote{rudeness}}, \emph{\enquote{flirting}} and \emph{\enquote{usher}}, while the farther words contained a range of aspects relevant to people's lives such as \emph{\enquote{billiards}}, \emph{\enquote{bronchitis}} and \emph{\enquote{depression}}.  
C6 contained close to the cluster center simple work-related words, e.g. \emph{\enquote{working}}, \emph{\enquote{employed}} and \emph{\enquote{student}}, while farther from the center there were more words referring to family members and even further away became the words referring to specific professions such as \emph{\enquote{psychologists}}, \emph{\enquote{barrister}} and \emph{\enquote{chemist}}. The values of C6 feature for AD patients were significantly lower than for controls. Finally, the cluster C9 contained simple business-related words close to the cluster center, such as \emph{\enquote{manage}}, \emph{\enquote{product}} and \emph{\enquote{account}}, while the words got more specific farther away from the centroid, e.g. \emph{\enquote{licensed}}, \emph{\enquote{reorganisation}} and \emph{\enquote{textile}}. 

Also, we checked how many words were considered as ICUs (words with  $d_{scaled}< 3.0$ to their closest cluster center) on AMI data and found that most words were counted. This suggests that the automatically computed SID is in fact very close to the simple proportion of nouns and verbs in the transcripts. In order to check this, we extracted the normalised counts of nouns and verbs from all transcripts in both datasets and used it to train single feature logistic regression classifiers. We obtained the precision 67.6, recall 66.8 and F-score 66.6 on DementiaBank and precision 77.1, recall 76.0 and F-score 74.3 on AMI dataset. 
Also, we found that on DementiaBank the simple bag-of-words baseline obtained the results very close to the current state-of-the-art that uses much more complex feature sets, including both acoustic and lexicosyntactic features \cite{Fraser2015}. These two observations suggest that there is still room for studying simple feature sets for predicting AD.

\section{Conclusion}

We experimented with two different definitions of idea density---propositional idea density and semantic idea density---in the classification task for predicting Alzheimer's disease. In the AD and psycholinguistic literature, PID has been automatically calculated using CPIDR \cite{Engelman2010,Ferguson2014,Bryant2013,Moe2016}.  
We show that CPIDR has a number of flaws when applied to AD speech, and we propose a new PID computation method DEPID which is more highly correlated with manual estimates of PID.  
We recommend that AD researchers use our automatic measure, DEPID-R, which also excludes repeating ideas from the total idea count, in place of CPIDR. 

This is the first comparison between PID and SID and also the first computational study that evaluates the predictive models for Alzheimer's disease on two very different datasets.
While on the closed-topic picture description dataset SID performs better, including PID also adds a small improvement to the classification results.
On the open-domain dataset we found that the PID was more predictive than SID as expected. However, the small number of automatically extracted cluster features underlying the SID, modeling the broad discussion topics, led to even better results.

In future we plan to study the usefulness and applicability of both PID and SID also in other clinical tasks, such as in clinical diagnostic tasks for depression or schizophrenia. 
Another possible avenue for future work would include combining dependency-base PID and embedding-based SID into a unified idea density measure that would take into account both the propositional structure as well as the semantic content of words.

\section*{Acknowledgements}

This research was supported by a Google award
through the Natural Language Understanding Focused Program, and under the Australian Research Council's Discovery Projects funding scheme (project number DP160102156), and in part by funding to ForeFront, a collaborative research group dedicated to the study of frontotemporal dementia and motor neuron disease, from the National Health and Medical Research Council (NHMRC) (APP1037746), and the Australian Research Council (ARC) Centre of Excellence in Cognition and its Disorders Memory Program (CE11000102). OP is supported by an NHMRC Senior Research Fellowship (APP1103258). 

\bibliography{references}

\begin{thebibliography}{}
\expandafter\ifx\csname natexlab\endcsname\relax\def\natexlab#1{#1}\fi

\bibitem[{Ahmed et~al.(2013{\natexlab{a}})Ahmed, de~Jager, Haigh, and
  Garrard}]{Ahmed2013a}
Samrah Ahmed, Celeste~A. de~Jager, Anne-Marie Haigh, and Peter Garrard.
  2013{\natexlab{a}}.
\newblock {Semantic processing in connected speech at a uniformly early stage
  of autopsy-confirmed Alzheimer's disease.}
\newblock {\em Neuropsychology\/} 27(1):79--85.

\bibitem[{Ahmed et~al.(2013{\natexlab{b}})Ahmed, Haigh, de~Jager, and
  Garrard}]{Ahmed2013}
Samrah Ahmed, Anne-Marie~F. Haigh, Celeste~A. de~Jager, and Peter Garrard.
  2013{\natexlab{b}}.
\newblock {Connected speech as a marker of disease progression in
  autopsy-proven Alzheimer's disease.}
\newblock {\em Brain\/} 136(12):3727--3737.

\bibitem[{Bayles et~al.(1985)Bayles, Tomoeda, Kaszniak, Stern, and
  Eagans}]{Bayles1985}
Kathryn~A. Bayles, Cheryl~K. Tomoeda, Alfred~W. Kaszniak, Lawrence~Z. Stern,
  and Karen~K. Eagans. 1985.
\newblock {Verbal perseveration of dementia patients}.
\newblock {\em Brain and Language\/} 25(1):102--116.

\bibitem[{Bayles et~al.(2004)Bayles, Tomoeda, McKnight, Helm-Estabrooks, and
  Hawley}]{Bayles2004}
Kathryn~A. Bayles, Cheryl~K. Tomoeda, Patrick~E. McKnight, Nancy
  Helm-Estabrooks, and Josh~N. Hawley. 2004.
\newblock {Verbal perseveration in individuals with Alzheimer's disease}.
\newblock {\em Seminars in Speech and Language\/} 25(4):335--347.

\bibitem[{Becker et~al.(1994)Becker, Boller, Lopez, Saxton, and
  McGonigle}]{Becker1994}
James~T. Becker, Francois Boller, Oscar~L. Lopez, Judith Saxton, and Karen~L.
  McGonigle. 1994.
\newblock {The natural history of Alzheimer's disease. Description of study
  cohort and accuracy of diagnosis.}
\newblock {\em Archives of Neurology\/} 51(6):585--594.

\bibitem[{Brown et~al.(2008)Brown, Snodgrass, Kemper, Herman, and
  Covington}]{Brown2008}
Cati Brown, Tony Snodgrass, Susan~J. Kemper, Ruth Herman, and Michael~A.
  Covington. 2008.
\newblock {Automatic measurement of propositional idea density from
  part-of-speech tagging.}
\newblock {\em Behavior research methods\/} 40(2):540--545.

\bibitem[{Bryant et~al.(2013)Bryant, Spencer, Ferguson, Craig, Colyvas, and
  Worrall}]{Bryant2013}
Lucy Bryant, Elizabeth Spencer, Alison Ferguson, Hugh Craig, Kim Colyvas, and
  Linda Worrall. 2013.
\newblock {Propositional Idea Density in aphasic discourse}.
\newblock {\em Aphasiology\/} 27(8):992--1009.

\bibitem[{Chand et~al.(2010)Chand, Baynes, Bonnici, and Farias}]{Chand2010}
Vineeta Chand, Kathleen Baynes, Lisa~M. Bonnici, and Sarah~Tomaszewski Farias.
  2010.
\newblock {Analysis of Idea Density (AID): A Manual}.
\newblock Technical report, University of California at Davis.

\bibitem[{Chand et~al.(2012)Chand, Baynes, Bonnici, and Farias}]{Chand2012}
Vineeta Chand, Kathleen Baynes, Lisa~M. Bonnici, and Sarah~Tomaszewski Farias.
  2012.
\newblock {A rubric for extracting idea density from oral language samples.}
\newblock {\em Current Protocols in Neuroscience\/} 1.

\bibitem[{de~Marneffe and Manning(2008)}]{Marneffe2008}
Marie-Catherine de~Marneffe and Christopher~D. Manning. 2008.
\newblock {Stanford Dependencies manual.}
\newblock Technical report, Stanford University.

\bibitem[{Engelman et~al.(2010)Engelman, Agree, Meoni, and Klag}]{Engelman2010}
Michal Engelman, Emily~M. Agree, Lucy~A. Meoni, and Michael~J. Klag. 2010.
\newblock {Propositional density and cognitive function in later life: findings
  from the Precursors Study.}
\newblock {\em Journals of Gerontology - Series B Psychological Sciences and
  Social Sciences\/} 65(6):706--711.

\bibitem[{Ferguson et~al.(2014)Ferguson, Spencer, Craig, and
  Colyvas}]{Ferguson2014}
Alison Ferguson, Elizabeth Spencer, Hugh Craig, and Kim Colyvas. 2014.
\newblock {Propositional idea density in women's written language over the
  lifespan: computerized analysis.}
\newblock {\em Cortex\/} 55:107--121.

\bibitem[{Fraser et~al.(2015)Fraser, Meltzer, and Rudzicz}]{Fraser2015}
Kathleen~C. Fraser, Jed~A. Meltzer, and Frank Rudzicz. 2015.
\newblock {Linguistic Features Identify Alzheimer's Disease in Narrative
  Speech.}
\newblock {\em Journal of Alzheimer's disease\/} 49(2):407--422.

\bibitem[{Goodglass and Kaplan(1983)}]{Goodglass1983}
Harold Goodglass and Edith Kaplan. 1983.
\newblock {\em The Assessment of Aphasia and Related Disorders\/}.
\newblock Lea \& Febiger.

\bibitem[{Jarrold et~al.(2014)Jarrold, Peintner, Wilkins, Vergryi, Richey,
  Gorno-Tempini, and Ogar}]{Jarrold2014}
William Jarrold, Bart Peintner, David Wilkins, Dimitra Vergryi, Colleen Richey,
  Maria~Luisa Gorno-Tempini, and Jennifer Ogar. 2014.
\newblock {Aided diagnosis of dementia type through computer-based analysis of
  spontaneous speech}.
\newblock In {\em Proceedings of the Workshop on Computational Linguistics and
  Clinical Psychology: From Linguistic Signal to Clinical Reality\/}. pages
  27--37.

\bibitem[{Jarrold et~al.(2010)Jarrold, Peintner, Yeh, Krasnow, Javitz, and
  Swan}]{Jarrold2010}
William~L. Jarrold, Bart Peintner, Eric Yeh, Ruth Krasnow, Harold~S. Javitz,
  and Gary~E. Swan. 2010.
\newblock Language analytics for assessing brain health: Cognitive impairment,
  depression and pre-symptomatic alzheimer's disease.
\newblock In {\em Proceedings of the 2010 International Conference on Brain
  Informatics\/}. pages 299--307.

\bibitem[{Kemper et~al.(2001)Kemper, Marquis, and Thompson}]{Kemper2001}
Susan Kemper, Janet Marquis, and Marilyn Thompson. 2001.
\newblock {Longitudinal change in language production: effects of aging and
  dementia on grammatical complexity and propositional content.}
\newblock {\em Psychology and Aging\/} 16(4):600--614.

\bibitem[{Kintsch and Keenan(1973)}]{Kintsch1973}
Walter Kintsch and Janice Keenan. 1973.
\newblock {Reading rate and retention as a function of the number of
  propositions in the base structure of sentences}.
\newblock {\em Cognitive Psychology\/} 5(3):257--274.

\bibitem[{Li and Nenkova(2015)}]{Li2015}
Junyi~Jessy Li and Ani Nenkova. 2015.
\newblock Fast and accurate prediction of sentence specificity.
\newblock In {\em Proceedings of the Twenty-Ninth AAAI Conference on Artificial
  Intelligence\/}. pages 2281--2287.

\bibitem[{MacWhinney(2000)}]{MacWhinney2000}
Brian MacWhinney. 2000.
\newblock {\em The {CHILDES} Project: Tools for analyzing talk, 3rd
  edition.\/}.
\newblock Lawrence Erlbaum Associates.

\bibitem[{Moe et~al.(2016)Moe, Breitborde, Shakeel, Gallagher, and
  Docherty}]{Moe2016}
Aubrey~M. Moe, Nicholas J.~K. Breitborde, Mohammed~K. Shakeel, Colin~J.
  Gallagher, and Nancy~M. Docherty. 2016.
\newblock {Idea density in the life-stories of people with schizophrenia:
  Associations with narrative qualities and psychiatric symptoms.}
\newblock {\em Schizophrenia Research\/} 172(1):201--205.

\bibitem[{Nicholas et~al.(1985)Nicholas, Obler, Albert, and
  Helm-Estabrooks}]{Nicholas1985}
Marjorie Nicholas, Loraine~K. Obler, Martin~L. Albert, and Nancy
  Helm-Estabrooks. 1985.
\newblock {Empty Speech in Alzheimer's Disease and Fluent Aphasia}.
\newblock {\em Journal of Speech and Hearing Research\/} 28(3):405--410.

\bibitem[{Orimaye et~al.(2014)Orimaye, Wong, and Golden}]{Orimaye2014}
Sylvester~O. Orimaye, Jojo Sze-Meng Wong, and Karen~J. Golden. 2014.
\newblock {Learning Predictive Linguistic Features for Alzheimer's Disease and
  related Dementias using Verbal Utterances}.
\newblock In {\em Proceedings of the Workshop on Computational Linguistics and
  Clinical Psychology: From Linguistic Signal to Clinical Reality\/}. pages
  78--87.

\bibitem[{Rentoumi et~al.(2014)Rentoumi, Raoufian, Ahmed, de~Jager, and
  Garrard}]{Rentoumi2014}
Vassiliki Rentoumi, Ladan Raoufian, Samrah Ahmed, Celeste~A. de~Jager, and
  Peter Garrard. 2014.
\newblock {Features and machine learning classification of connected speech
  samples from patients with autopsy proven Alzheimer's disease with and
  without additional vascular pathology.}
\newblock {\em Journal of Alzheimer's disease\/} 42(S3):3--17.

\bibitem[{Roark et~al.(2011)Roark, Mitchell, Hosom, Hollingshead, and
  Kaye}]{Roark2011}
Brian Roark, Margaret Mitchell, John-Paul Hosom, Kristy Hollingshead, and
  Jeffrey Kaye. 2011.
\newblock {Spoken Language Derived Measures for Detecting Mild Cognitive
  Impairment.}
\newblock {\em IEEE transactions on audio, speech, and language processing\/}
  19(7):2081--2090.

\bibitem[{Snowdon et~al.(1996)Snowdon, Kemper, Mortimer, Greiner, Wekstein, and
  Markesbery}]{Snowdon1996}
David~A. Snowdon, Susan~J. Kemper, James~A. Mortimer, Lydia~H. Greiner,
  David~R. Wekstein, and William~R. Markesbery. 1996.
\newblock {Linguistic ability in early life and cognitive function and
  Alzheimer's disease in late life. Findings from the Nun Study.}
\newblock {\em JAMA\/} 275(7):528--532.

\bibitem[{Tausczik and Pennebaker(2010)}]{Tausczik2010}
Yla~R. Tausczik and James~W. Pennebaker. 2010.
\newblock {The Psychological Meaning of Words: LIWC and Computerized Text
  Analysis Methods}.
\newblock {\em Journal of Language and Social Psychology\/} 29(1):24--54.

\bibitem[{Thomas et~al.(2005)Thomas, Keselj, Cercone, Rockwood, and
  Asp}]{Thomas2005}
Calvin Thomas, Vlado Keselj, Nick Cercone, Kenneth Rockwood, and Elissa Asp.
  2005.
\newblock {Automatic detection and rating of dementia of Alzheimer type through
  lexical analysis of spontaneous speech}.
\newblock In {\em IEEE International Conference Mechatronics and Automation,
  2005\/}. pages 1569--1574.

\bibitem[{Tomoeda et~al.(1996)Tomoeda, Bayles, Trosset, Azuma, and
  McGeagh}]{Tomoeda1996}
Cheryl~K. Tomoeda, Kathryn~A. Bayles, Michael~W. Trosset, Tamiko Azuma, and
  Anna McGeagh. 1996.
\newblock {Cross-sectional analysis of Alzheimer disease effects on oral
  discourse in a picture description task}.
\newblock {\em Alzheimer Disease and Associated Disorders\/} 10(4):204--215.

\bibitem[{Turner and Greene(1977)}]{Turner1977}
Althea Turner and Edith Greene. 1977.
\newblock {The construction and use of a propositional text base}.
\newblock Technical report, University of Colorado.

\bibitem[{Yancheva and Rudzicz(2016)}]{Yancheva2016}
Maria Yancheva and Frank Rudzicz. 2016.
\newblock {Vector-space topic models for detecting Alzheimer's disease}.
\newblock In {\em Proceedings of the 54th Annual Meeting of the Association for
  Computational Linguistics\/}. pages 2337--2346.

\end{thebibliography}
\bibliographystyle{acl_natbib}

\end{document}